\relax
\pdfoutput=1
\documentclass[letterpaper]{article} 
\usepackage{aaai20}  
\usepackage{times}  
\usepackage{helvet}  
\usepackage{courier}  
\usepackage{url}  
\usepackage{graphicx}  
\usepackage{amssymb}
\usepackage{amsmath}
\usepackage{booktabs}
\usepackage{tikz}

\newcommand{\cons}[1]{\texttt{#1}}

\definecolor{zerocolor}{RGB}{106,93,78}    
\definecolor{onecolor}{RGB}{86,62,46}      
\definecolor{twocolor}{RGB}{227,108,69}    
\definecolor{threecolor}{RGB}{235,166,81}  
\definecolor{fourcolor}{RGB}{120,160,92}   
\definecolor{fivecolor}{RGB}{109,139,126}  
\definecolor{sixcolor}{RGB}{122,130,163}   
\definecolor{sevencolor}{RGB}{175,126,138} 
\definecolor{eightcolor}{RGB}{234,140,144} 
\definecolor{ninecolor}{RGB}{216,109,97}   

\newcommand{\drawzero}[1]{\draw [fill,color=zerocolor] (#1) -- +(3cm,0cm) -- +(3cm,4cm) -- +(0cm,4cm) -- +(0cm,0cm) ++(1cm,1cm) -- +(1cm,0cm) -- +(1cm,2cm) -- +(0cm,2cm) -- cycle;}
\newcommand{\drawone}[1]{\draw [fill,color=onecolor] (#1)++(1cm,0cm) -- +(1cm,0cm) -- +(1cm,4cm) -- +(-1cm,4cm) -- +(-1cm,3cm) -- +(0cm,3cm) -- cycle;}
\newcommand{\drawtwo}[1]{\draw [fill,color=twocolor] (#1)-- +(3cm,0cm) -- +(3cm,1cm) -- +(2cm,1cm) -- +(2cm,2cm) -- +(3cm,2cm) -- +(3cm,4cm) -- +(1cm,4cm) -- +(1cm,2cm) -- +(0cm,2cm) -- cycle;}
\newcommand{\drawthree}[1]{\draw [fill,color=threecolor] (#1)-- +(3cm,0cm) -- +(3cm,4cm) -- +(0cm,4cm) -- +(0cm,3cm) -- +(2cm,3cm) -- +(2cm,2cm) -- +(1cm,2cm) -- +(1cm,1cm) -- +(0cm,1cm) -- cycle;}
\newcommand{\drawfour}[1]{\draw [fill,color=fourcolor] (#1)++(1cm,0cm) -- +(2cm,0cm) -- +(2cm,2cm) -- +(1cm,2cm) -- +(1cm,3cm) -- +(2cm,3cm) -- +(2cm,4cm) -- +(0cm,4cm) -- +(0cm,2cm) -- +(-1cm,2cm) -- +(-1cm,1cm) -- +(0cm,1cm) -- cycle;}
\newcommand{\drawfive}[1]{\draw [fill,color=fivecolor] (#1)-- +(3cm,0cm) -- +(3cm,4cm) -- +(0cm,4cm) -- +(0cm,2cm) -- +(2cm,2cm) -- +(2cm,1cm) -- +(0cm,1cm) -- cycle;}
\newcommand{\drawsix}[1]{\draw [fill,color=sixcolor] (#1)-- +(3cm,0cm) -- +(3cm,2cm) -- +(1cm,2cm) -- +(1cm,3cm) -- +(2cm,3cm) -- +(2cm,4cm) -- +(0cm,4cm) -- cycle;}
\newcommand{\drawseven}[1]{\draw [fill,color=sevencolor] (#1)-- +(1cm,0cm) -- +(1cm,1cm) -- +(2cm,1cm) -- +(2cm,3cm) -- +(3cm,3cm) -- +(3cm,4cm) -- +(0cm,4cm) -- +(0cm,3cm) -- +(1cm,3cm) -- +(1cm,2cm) -- +(0cm,2cm) -- cycle;}
\newcommand{\draweight}[1]{\draw [fill,color=eightcolor] (#1)-- +(2cm,0cm) -- +(2cm,2cm) -- +(3cm,2cm) -- +(3cm,4cm) -- +(1cm,4cm) -- +(1cm,2cm) -- +(0cm,2cm) -- cycle;}
\newcommand{\drawnine}[1]{\draw [fill,color=ninecolor] (#1)-- +(2cm,0cm) -- +(2cm,2cm) -- +(3cm,2cm) -- +(3cm,4cm) -- +(0cm,4cm) -- cycle;}

\def\orcidID#1{\unskip$^{[#1]}$}

\nocopyright

\allowdisplaybreaks[1]

\begin{document}

\title{Nmbr9 as a Constraint Programming Challenge}
\author{Mikael Zayenz Lagerkvist\orcidID{0000-0003-2451-4834}\\
  \url{research@zayenz.se}, \url{https://zayenz.se}}

\maketitle

\section{Introduction}
\label{sec:introduction}

Modern board games are a rich source of interesting and new challenges
for combinatorial problems. The game Nmbr9 is a solitaire style puzzle
game using polyominoes. The rules of the game are simple to explain,
but modelling the game effectively using constraint programming is
hard.

This abstract presents the game, contributes new generalized variants
of the game suitable for benchmarking and testing, and describes a
model for the presented variants. The question of the top possible
score in the standard game is an open challenge.

\section{Nmbr9}
\label{sec:nmbr9}

\begin{figure}
\centering
\begin{tikzpicture}[even odd rule,scale=0.32]
  \drawzero{0cm,6cm}
  \drawone{4cm,6cm}
  \drawtwo{8cm,6cm}
  \drawthree{12cm,6cm}
  \drawfour{16cm,6cm}
  \drawfive{0cm,0cm}
  \drawsix{4cm,0cm}
  \drawseven{8cm,0cm}
  \draweight{12cm,0cm}
  \drawnine{16cm,0cm}
\end{tikzpicture}
\caption{Parts to place representing the numbers 0-9.}
    \label{fig:parts}
\end{figure}
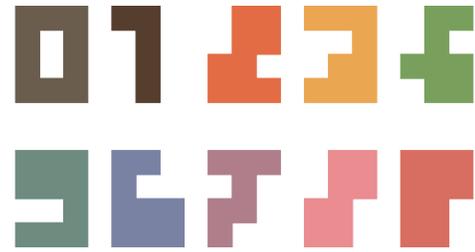

Nmbr9~\cite{game:nmbr9} is a solitaire puzzle game using polyominoes
that can be played by multiple people at the same time. The goal of
the game is to build layers of the polyominoes in
Figure~\ref{fig:parts} according to a pre-defined but unknown shuffle,
and to place polyominoes representing larger numbers higher up to
score points. 

\subsection{Rules}
\label{sec:nmbr9:rules}

The game consists of polyominoes representing the numbers 0 to 9, two
copies per number for each player. A common deck of 20 cards with the
polyominoes are shuffled.

When a card is drawn, each player takes the corresponding part and
places it in their own area. Placements are done in levels, where the
first level (level $1$) is directly on the table, and level $n$ is on top of
level $n-1$.

Placement must always be done fully supported (level $1$ is always
fully supported, level $n$ is supported by previously placed parts in level
$n-1$). Each level must be fully 4-connected (i.e., not diagonally)
after placement. The final requirement on placements is that when placing a
part on top of other parts, it must be on top of at least two
different parts. For example, if a 9-part is in level $n$, then an
8-part can not be placed in level $n+1$ resting solely on the 9-part.

When all cards have been drawn, the score is computed as the the sum
of the value for parts times their level minus one. For example, an 8-part on
level 3 is worth 16, while it is worth 0 on level 1.

\subsection{Variants}
\label{sec:nmbr9:variants}

We define variants of Nmbr9 to get a larger sampling of problems to
solve, from small trivial problems to larger intractable problems. Let
$T{-}m{-}c{-}n$ be the scheme, where
\begin{description}
\item[$T\in{U,F,K}$] Whether the draft is \emph{unknown}, \emph{free} to choose,
  or \emph{known}. The first corresponds to a quantified problem, where the
  draft is $\forall$-quantified. The second models the question of the
  max possible score, while the last models the max possible score
  given a specific known shuffle.
\item[$m$] The maximum value to use, from 0-9. 
\item[$c$] The number of available copies of each polyomino.
\item[$k$] The number of cards in the deck to use, $k\leq m\cdot c$
  must hold.
\end{description}

With this taxonomy of variants, the standard game is $U{-}9{-}2{-}20$, where
all available parts are used. Reducing $m$, $c$, or $k$ are all
effective ways to make the model smaller.  $F$ variants are
interesting for answering the question of the top-score possible. $K$
variants are interesting computationally because they represent a much
simpler but still hard problem, and are also interesting
recreationally given a finished game, where participants wonder how
they compare to the best possible result for that particular shuffle.

At Board Game Geek, a forum thread on solving the standard
$F{-}9{-}2{-}20$ instance has a top
score of 229 points, using a total of 7 layers~\cite{Kuenzler2018}.

\section{Constraint programming model}
\label{sec:model}

In this section, a constraint programming model is described for Nmbr9
in the $F$ and $K$ variants; a quantified model for the $U$ variant
could be built as an extension on top of this. The model is based on
the models in~\cite{Lagerkvist2008,Lagerkvist2019}, where
\cons{regular} constraints are used for placement of polyominoes in
grids. 

Nmbr9 is played on a conceptually infinite grid, with as many layers
as needed available. The model used here requires a fixed grid. In
practice, placements very seldom surpass 20 squares wide, and more
than 7 layers is unlikely. The width/height size $s$ and the number of
layers $l_\top$ are parameters for the model. A core problem is
keeping levels connected. This is accomplished using the technique of
defining an area around parts, here as a new value. Also, parts may
not be placed, which is controlled using reification of the regular
expression. As an example, the regular expression for part 0 (called
$R_0$) in its two rotations, where the value 0 is an empty square, 1
is a square occupied by the part, and 2 is an area surrounding the
part becomes (including initial control variable for reified placement)
\begin{align*}
  10^*(&2^30^{s-4}21^320^{s-5}(210120^{s-5})^221^320^{s-4}2^3\,|\\
  &2^40^{s-5}21^420^{s-6}210^2120^{s-6}21^420^{s-5}2^4)0^*\,|\,00^*
\end{align*}

Let $P=\{p_1,\ldots,p_n\}$ ($n=m\cdot c$) be the set of parts, and
$v(p)\in P\to {0..m}$ the value of each part. Variables
$D=\langle d_1,\ldots,d_k\rangle\in P$ represent the deck of
polyominoes, and $O=\langle o_1,\ldots,o_n\rangle\in\{1..n\}$ the
order of the polyominoes as they occur in the deck, with values above
$k$ representing not used. The matrix $G_l$ is the grid for level $l$
for all parts on that level (of size $s\times s$, domain $0..n$,
border is 0), and $G_{lp}$ the grid for a part $p$ on level $l$ (of
similar size, domain $0..2$). Variables $G_{lp}^1$ and $G_{lp}^2$ are
Boolean matrices representing when $G_{lp}$ is 1 and 2
respectively. Boolean variables $L_{pl}$ represent $p$ being placed on
level $l$, integer variable $L_p\in {0..l_ \top}$ the level of $p$ (if
any, 0 if not), and $Y_p$ a Boolean variable representing that $p$ is
placed with $N_p$ the inverse. The following defines the constraints
of the model, where $p, p'$ is implicitly assumed to range over $P$
with $p\neq p'$, $l$ over $1..l_\top$, $l'$ over $2..l_\top$, and
$i,j$ over $1..s$. Operators $\dot{\lor}$ and $\dot{\land}$ are used
for point-wise logical operations. The notation $[M]$ is used to
indicate true iff any element in the matrix M is true.

\begin{gather}
  \cons{global\_cardinality}(D, \langle
  Y_1,\ldots,Y_n\rangle) \label{cons:gcc}\\
  \cons{regular}(R_p, L_{pl}G_{lp})\label{cons:placement}\\
  \cons{inverse}(O, \langle D_1,\ldots,D_k,E_{k+1},\ldots,E_n\rangle),
  E\in P\label{cons:orderchannel}\\
  B(p,p') \leftrightarrow O(p) < O(p'),\, O(p)\leq k\leftrightarrow Y_p \label{cons:orderchannel2}\\
  \cons{int\_to\_bool}(L_p, \langle N_p, L_{p1},\ldots,L_{pl_\top}\rangle), \label{cons:levelchannel}\\
  Y_p = 1 - N_p,
  \label{cons:minorchannel}\\
  G_{lp}(i,j)=1\leftrightarrow G_{lp}^1(i,j), G_{lp}(i,j)=2\leftrightarrow G_{lp}^2(i,j)  \label{cons:aspectchannel}\\
  G_l(i,j) = p \leftrightarrow G_{lp}(i,j) = 1 \label{cons:gridchannel}\\
  \left(L_{pl}\land\exists_{p''|B(p'', p)}L_{p''l}\right)\rightarrow \left[G_{lp}^2 \dot{\land} \left(\dot{\lor}_{p'|B(p',p)}
      G_{lp'}^1\right)\right]
  \label{cons:connected}\\
  G_{l'p}^1(i,j) \rightarrow \lor_{p'|B(p',p)} G_{l'-1p'}^1(i,j)
  \label{cons:ontop}\\
  L_{pl'}\rightarrow\left(2\leq\textstyle \sum_{p'|B(p',p)} [G_{lp}^1 \dot{\land} G_{l-1p'}^1]\right)
  \label{cons:atleasttwo}\\
  S = \textstyle \sum_{p\in P} (L_p-Y_p)\cdot v(p)\label{cons:score}
\end{gather}

Constraint~\ref{cons:gcc} checks that the deck is a shuffle of
parts. Constraint~\ref{cons:placement} is the core placement
constraints, with $l_\top\cdot n$ regular constraints in
total. Constraints~\ref{cons:orderchannel} to~\ref{cons:gridchannel}
channel information between variables.
Constraints~\ref{cons:connected} to~\ref{cons:atleasttwo} implement
the requirements for connectedness and being placed supported and on
top of at least two different parts. When $k=n$, then
constraint~\ref{cons:gcc} is equivalent to an \cons{alldifferent}, and
in constraint~\ref{cons:orderchannel} there are no extra $E$ variables
representing fake deck placement for non-used parts.

Finally, constraint~\ref{cons:score} sums up the score of the
solution, where $L_p-Y_p$ is 0 for non-used parts ($L_p=0$ and
$Y_p=0$), and also for parts in the first layer ($L_p=1$ and $Y_p=1$).

\section{Implementation}
\label{sec:implementation}

The described model has been implemented using Gecode~\cite{gecode}
version 6.2.0 and C++17, and can be accessed at
\url{github.com/zayenz/cp-2019-nmbr9}.

Choices are made on the deck, then the levels of parts, then the
placements. A static variable order is used spiralling out from the
centre to keep placement centered in the grid for as long as possible.
Extra implied constraints that the first card is on the first level,
and that the area for each level is at most the area for the level
above are added. The latter improves performance significantly.

It is intractable to find a max score
for the standard game approaching the score found manually using this
model and search strategy; the model has more than 2.5 million
propagators, and setting up the root node takes 5 seconds. A small
problem like $F{-}6{-}1{-}5$ with 3 levels and grid size 8 takes more
than 30 seconds and 80k failures to solve.

\section{Conclusions}
\label{sec:conclusions}

As seen, modelling Nmbr9 is surprisingly complex, 
mainly due to the connectedness and support
requirements. This leads to a large model with not much
propagation. Finding the maximum score for the standard game is a
computationally hard open challenge.

Future work include: exploring how large problems can be solved with
the current model; investigating better models for stronger
propagation; and finding effective heuristics for the search. For
search heuristics, the \emph{propagation guided global regret}
in~\cite{Lagerkvist2019} might be very interesting. 

\paragraph*{ Acknowledgments}
Thanks to Magnus Gedda, Magnus Rattfeldt, and Martin Butler for
interesting and informative discussions on games and search methods.

\bibliographystyle{aaai}
\fontsize{9.0pt}{10.0pt}
\bibliography{references}

\begin{thebibliography}{}

\bibitem[\protect\citeauthoryear{{Gecode team}}{2019}]{gecode}
{Gecode team}.
\newblock 2019.
\newblock {G}ecode, the generic constraint development environment.
\newblock \url{http://www.gecode.org/}.

\bibitem[\protect\citeauthoryear{Kuenzler}{2018}]{Kuenzler2018}
Kuenzler, P.
\newblock 2018.
\newblock {Board Game Geek} -- {Nmbr9} {Forums, What is the highest possible
  score?}
\newblock \url{https://boardgamegeek.com/article/30607221#30607221}, [Accessed
  2019-08-27].

\bibitem[\protect\citeauthoryear{Lagerkvist and Pesant}{2008}]{Lagerkvist2008}
Lagerkvist, M.~Z., and Pesant, G.
\newblock 2008.
\newblock Modeling irregular shape placement problems with regular constraints.
\newblock In {\em First Workshop on Bin Packing and Placement Constraints
  BPPC'08}.

\bibitem[\protect\citeauthoryear{Lagerkvist}{2019}]{Lagerkvist2019}
Lagerkvist, M.~Z.
\newblock 2019.
\newblock State representation and polyomino placement for the game patchwork.
\newblock In {\em The 18th workshop on Constraint Modelling and Reformulation}.

\bibitem[\protect\citeauthoryear{Wichmanm}{2017}]{game:nmbr9}
Wichmanm, P.
\newblock 2017.
\newblock {NMBR} 9.

\end{thebibliography}

\end{document}